\documentclass{article}
\usepackage{spconf}

\usepackage{graphicx}
\usepackage[hyphens]{url} 
\usepackage[pagebackref=true,breaklinks=true,letterpaper=true,colorlinks,bookmarks=false]{hyperref}
\usepackage{float}

\usepackage{amsmath}
\usepackage{amssymb}

\usepackage[ruled]{algorithm2e}
\usepackage[noend]{algpseudocode}
\usepackage{lipsum}
\usepackage{filecontents}
\usepackage{url}
\usepackage[para]{footmisc}
\usepackage{multirow}

\title{A Feature Embedding Strategy for High-level CNN representations from Multiple ConvNets}

\twoauthors
   {T. Akilan, Q.M. Jonathan Wu\thanks{This work was made possible by the facilities of the Shared Hierarchical Academic Research Computing Network (SHARCNET:www.sharcnet.ca) and Compute/Calcul Canada.}}
	{Department of Electrical and \\Computer Engineering\\
    University of Windsor\\
    Canada}
  {Wei Jiang}
	{Department of Control Science and \\Engineering\\
	Zhejiang University\\
	China}

\begin{document}
%
\maketitle
%

\begin{abstract}
Following the rapidly growing digital image usage, automatic image categorization has become preeminent research area. It has broaden and adopted many algorithms from time to time, whereby multi-feature (generally, hand-engineered features) based image characterization comes handy to improve accuracy. Recently, in machine learning, pre-trained deep convolutional neural networks (DCNNs or ConvNets) have proven that the features extracted through such DCNN can improve classification accuracy. Thence, in this paper, we further investigate a feature embedding strategy to exploit cues from multiple DCNNs. We derive a generalized feature space by embedding three different DCNN bottleneck features with weights respect to their softmax cross-entropy loss. Test outcomes on six different object classification data-sets and an action classification data-set show that regardless of variation in image statistics and tasks the proposed multi-DCNN bottleneck feature fusion is well suited to image classification tasks and an effective complement of DCNN. The comparisons to existing fusion-based image classification approaches prove that the proposed method surmounts the state-of-the-art methods and produces competitive results with fully trained DCNNs as well.
\end{abstract}


\section{Introduction}

The traditional classification models using single feature representation suffers from the inability to tackle intra-class variations and global variants such as color, lightings and orientation of image statistics. Therefore, it is an intuitive process to fuse multiple features to meliorate the classification accuracy because multiple features can plausibly create a well generalized feature space. Researchers in the computer vision community also have shown interest in multiple feature fusion.

For example, Li \emph{et al.} \cite{Li08} utilized the Riemann manifold to combine the features from the covariance matrix of multiple features and concatenated multiple features to represent the object appearance. Meanwhile, Park \cite{Park10} took the Multi-partitioned feature-based classifier (MPFC) to fuse features such as Hue-saturation-value(HSV), Discrete cosine transformation (DCT) coefficients, Wavelet packet transform (WPT) and Hough transform (HT) with specific decision characteristic expertise table of local classifiers. Similarly, Kwon \emph{et al.} \cite{Kwon10} had advantage of multiple features for efficient object tracking, where, they dissevered the task into multiple constituents and combined multiple features through sparse Principal component analysis (PCA) to select the most important features, by which, the appearance variations were captured.

On the other hand, researchers in \cite{Fernando12}, \cite{Gehler09}, \cite{Khan12}, \cite{Dixit15} also found different ways to merge multiple hand-engineered-features to improve classification accuracy. Fernando \emph{et al.} \cite{Fernando12} merged Hue-histograms, Color name (CN) descriptors, Scale-invariant feature transform (SIFT) and Color-SIFT, while, Gehler and Nowozin \cite{Gehler09} achieved some success of improving classification accuracy by means of combining the basic SIFT feature with another eight different features: Histogram of gradients (HOG), Local binary pattern (LBP), Color-SIFT and so forth using Multiple kernel learning (MKL) to combine 49 different kernel matrices. Khan \emph{et al.} \cite{Khan12} employed multiple cues by individually processing shape and color cues then combining them by modulating the SIFT shape features with category-specific color attention. They used a standardized multi-scale grid detector with Harris-laplace point detector and a blob detector to create feature description, then they normalized all the patches to a predefined size and computed descriptors for all regions. Dixit \emph{et al.} \cite{Dixit15} embedded features from a CNN with Semantic fisher vector (SFV), where the  SFV is ciphered as parameters of a multi-nominal Gaussian mixture FV.

In the aforesaid literature, however, the features fused are mainly the hand-engineered features or such features with bottleneck features\footnote{The high-level feature representations of ConvNet that is feed into a final classification layer is called bottleneck features.} from a single CNN. Hence, utilizing the bottleneck features extracted through an off-the-shelf pre-trained CNN, significantly, outperforms a majority of the baselines state-of-the-art methods \cite{Razavian_2014_CVPR_Workshops}.
Thus, one may ponder the following questions: (i) If multiple CNN features extracted from different networks, can such features be complementary?, if so (ii)  what can be an acceptable approach to fuse them so that the classification accuracy will improve? We address these questions by carrying out experiments on various data-sets with three different pre-trained CNNs as feature extractors, weights based on cross-entropy loss function as feature embedding scheme and softmax as classifier. The experiment results have strengthen our idea of fusing multiple CNN features to improve image classification accuracy.

\subsection{CNN as Feature Extractor}\label{cnn_featrue_extractor}

A DCNN pre-trained on large image data-set can be exploited as generic feature extractor through transfer learning process \cite{Oquab:2014:LTM:2679600.2680210}. Generally, in transfer learning, parameters (weights and biases) of first $n$ layers of source (pre-trained DCNN) are transferred to the first $n$ layers of target (new task) network and left without updates during training on new data-set, while the rest of the layers known as adaptation layers of target task are randomly initialized and updated over the training. If a fine-tuning strategy is taken then back-propagation process will be carried out through the entire (copied + randomly initialized layers) network  for calibrating the parameters of the copied layers in the new network so that the DCNN responses well to the new task.

In this experiment, we take three pre-trained networks: AlexNet, VGG-16, and Inception-v3 and extract features from their respective penultimate layers. These networks have been trained on ImageNet\footnote{It contains more than 14 million images which are hand labeled with the presence/absence of 21000+ categories.}, where the final logits layer of each network has 1000 output neurons. That final layer is decapitated, then rest of the DCNN is employed as fixed feature extractor on the new data-sets, where number classes per data-set may differ. The following intermezzo highlights the properties of the DCNNs.

\textbf{AlexNet}\cite{Krizhevsky12} is the winner of 2012 ImageNet Large Scale Visual Recognition Challenge (ILSVRC) with 37.5\% and 17.0\% top-1 and top-5 object classification error rates respectively. It subsumes 5 convolutional (Conv) layers occasionally interspersed with max-pooling layers, 3 fully-connected (FC) layers and the last softmax classifier with 1000 output neurons trained on 1.2 million images in the ImageNet-2010 data-set. The penultimate layer referred as \emph{FC7} has 4096 output channels.
\textbf{VGG-16}\cite{DBLP:journals/corr/SimonyanZ14a} is the winner of 2014 ILSVRC challenge for localization task with 25.3\% error and runner-up of the classification task with 24.8\% and 7.5\% top-1 and top-5 error rates respectively. It has 16 Conv layers with maxpooling layers after each set of two or more Conv layers, 2 FC layers, and a final softmax output layer. The penultimate layer \emph{FC2} has 4096 channels of output.
\textbf{Inception-v3}\cite{DBLP:journals/corr/SzegedyLJSRAEVR14} is an improved version of GoogLeNet the winner of 2014 ILSVRC classification task. It achieved 21.2\% top-1 and 5.6\% top-5 error rates on the benchmark ILSVRC 2012 classification challenge validation set. We extract features of target data-sets from a maxpooling layer named as \emph{pool\_3:0} in the network, which has 2048 output channels.

Rest of this paper is organized as follows. Section \ref{system_overview} expatiates on the main ideas: feature extraction, feature embedding and classification via block diagrams and mathematical derivations. Section \ref{experiments} details the experimental results through quantitative and qualitative analysis. Finally, Section \ref{conclusion} concludes the work with final remarks on future directions.

\section{System Overview}\label{system_overview}

As described in Section \ref{cnn_featrue_extractor}, using the selected CNN models and their associated learned parameters a forward-pass operation (without back-propagation) is carried out on the image statistics of new data-sets to extract bottleneck features. Depends on the size of the data-set, feature extraction process may take several hours; however, it will be considerably little time than training or fine-tuning the CNN completely. For instance, on a Intel(R) Core(TM) i7-3770 CPU @ 3.40GHz machine with 16.0GB RAM, it would take about 5-6 hours to get the features from CIFAR10 data-set through Inception-v3.

\begin{figure}[!htp]
  \begin{center}
    \includegraphics[width=85mm]{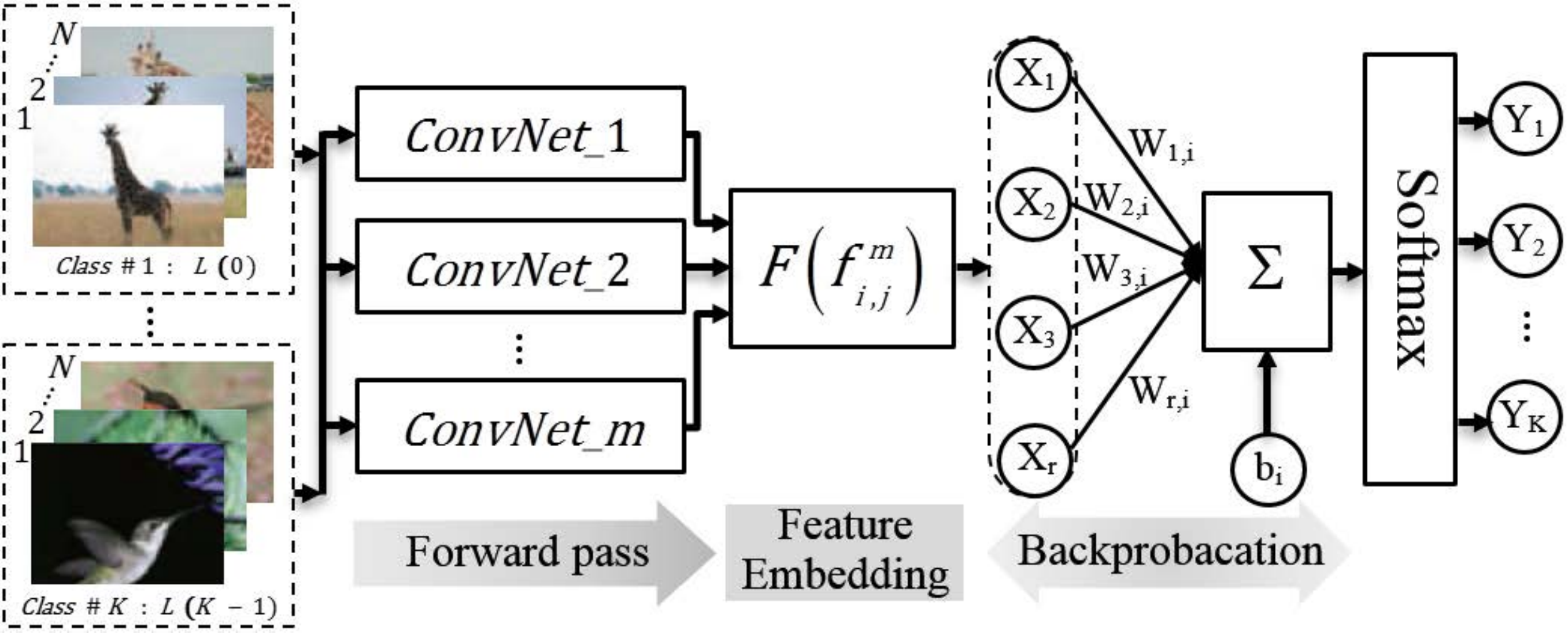}
  \end{center}
  \caption{The Image Classification System Overview (ConvNet refers to CNN).}
  \label{fig:over_all_system}
\end{figure}


\subsection{Feature Embedding}
\begin{figure}[!htp]
  \begin{center}
    \includegraphics[width=85mm]{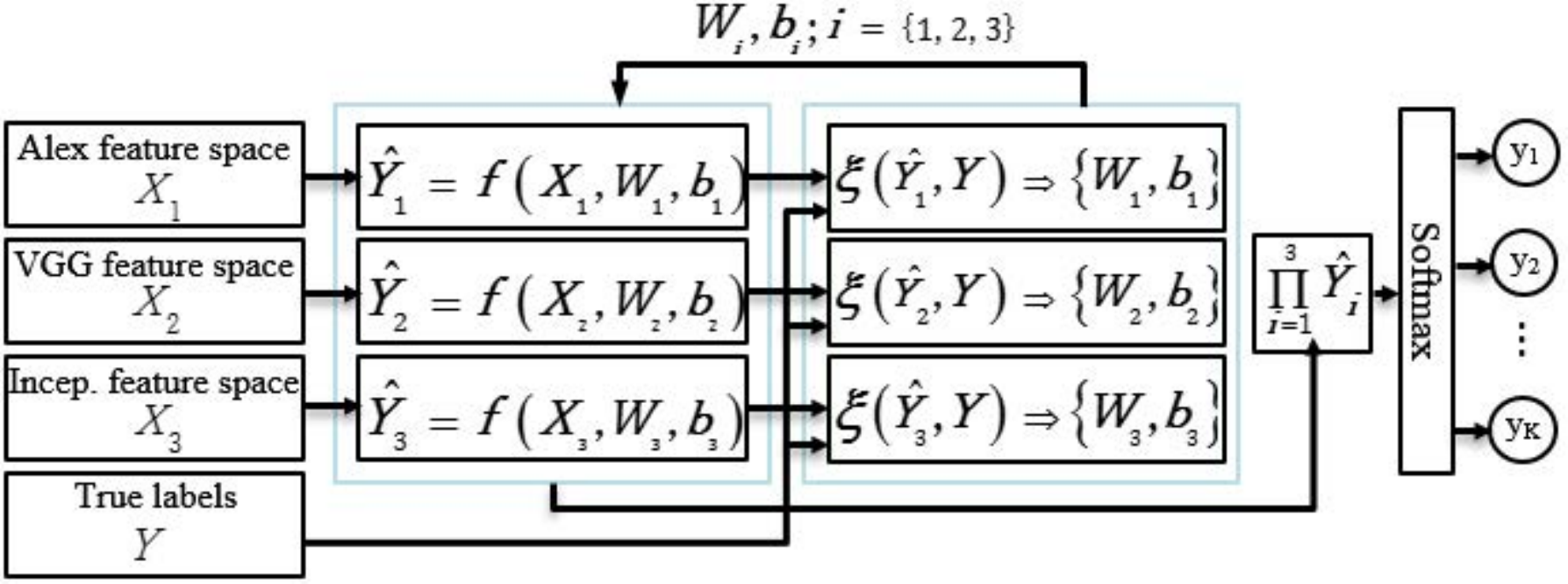}
  \end{center}
  \caption{Weighted Cross-entropy Based Feature Embedding.}
  \label{fig:feature_embedding}
\end{figure}

As we exploit three different CNNs for feature extraction as shown in Figure~\ref{fig:over_all_system}, the system must be devised with an appropriate strategy to merge the extracted features toward classification accuracy gain. The basic approach is concatenating all different features in a single vector per sample as $F=\{f_1^{n\times p1}, f_2^{n\times p2},\cdots f_m^{n\times pm}\}$, thus the final feature space $F$ has the dimension of $n\times (p1+p2+\cdots+pm)$. Although, such straight forward concatenation process often improve classification accuracy than using single feature, the penalty is unfair since a weak feature may deteriorate the performance of other good features. We circumvent that by introducing weighted feature embedding layer as shown in Figure~\ref{fig:feature_embedding}, where we calculate cross-entropy loss for each feature individually and update their assigned parameters using softmax function and gradient descent based optimizer to minimize the cross-entropy loss. On the other hand, this layer functions as indemnifier for the variant image statistics like imaging conditions, viewpoints and object types of the source and target data. The following snippet describes the mathematical background of the technique.

The softmax function produces a categorical probability distribution, when the input is a set of multi-class logits as:
\begin{equation} \label{eqn:softmax}
\resizebox{0.25\textwidth}{!}{$
\sigma(z)_j = \frac{e^{z_j}}{\sum^K_{k=1}e^{z_j}} for j=1,...,K,$}
\end{equation}
where input $Z$ is $K$-dimensional vector and output is also a $K$-dimensional vector having real values in the range $(0,1)$ and that add up to 1 as normalization happens via the sum of exponents term dividing actual exponentiation term. The cost function for the softmax function of the model can be written in terms of likelihood maximization with a given set of parameter $\varphi$ as:
\begin{equation} \label{eqn:likelihood}
\resizebox{0.15\textwidth}{!}{$
\mathop {\arg \max }\limits_\varphi  \mathcal{L} (\varphi|,\bf{t},\bf{z}),$}
\end{equation}
where the likelihood can be deduced to a conditional distribution of $\bf{t}$ and $\bf{z}$ for the same $\varphi$ as:
\begin{equation} \label{eqn:conditional_probability}
\resizebox{0.20\textwidth}{!}{$
P(t, z|\varphi) = P(t|z,\varphi)P(z|\varphi)$}.
\end{equation}
Note that the probability that the class $t=j$ for a given input $z$ and with $j=1,...,K$ can be written in matrix form as:
\begin{equation} \label{eqn:class_probability}
\resizebox{0.425\textwidth}{!}{$
\left[ {\begin{array}{*{20}{c}}
{P(t = 1|{\bf{z}})}\\
 \vdots \\
{P(t = K|{\bf{z}})}
\end{array}} \right] = \left[ {\begin{array}{*{20}{c}}
{\sigma {{({\bf{z}})}_1}}\\
 \vdots \\
{\sigma {{({\bf{z}})}_K}}
\end{array}} \right] = \frac{1}{{\sum\limits_{j = 1}^K {{e^{{z_j}}}} }}\left[ {\begin{array}{*{20}{c}}
{{e^{{z_1}}}}\\
 \vdots \\
{{e^{{z_K}}}}
\end{array}} \right],$}
\end{equation}
where $P(\bf{t},j|\bf{z})$  is the probability that the class is $j$ given that the input is $z$. Eventually, the cost function through maximizing the likelihood can be done by minimizing the negative log-likelihood as:
\begin{equation} \label{eqn:cost_fun}
\resizebox{0.425\textwidth}{!}{$- log{\cal L}(\theta |{\bf{t}},{\bf{z}}) = \xi ({\bf{t}},{\bf{z}}) =  - log\prod\limits_{j = 1}^K {y_j^{{t_j}}}  =  - \sum\limits_{j = 1}^K {{t_j}}  \cdot log({y_j})$},
\end{equation}
where $\xi$ denotes the cross-entropy error function. Then, the derivative $\partial \xi / \partial {W}$ of the cost function with respect to the softmax input $z$ can be used to update the weights as:
\begin{equation}\label{equn:weight_updates}
\resizebox{0.20\textwidth}{!}{$
W(t+1) = W(t) - \lambda \frac{{\partial \xi }}{{\partial w(t)}}$},
\end{equation}
where $\lambda$ the learning rate tells us how quickly the cost changes the weights. In the same way, biases can also be updated; towards the goal of bringing the error function to local minimum. In this work, we utilize the backpropagation (aka backprops) based on gradient descendant optimization algorithm to update the weights and biases. The gradient decent algorithm is the workhorse of learning in neural networks, these days. Intricate description of backprops can be referred from \cite{Michael15}. Thus, we get dimension reduced logits $\hat Y_1, \hat Y_2, \hat Y_3$ of the Alex, VGG, and Inception bottleneck features respectively as shown in Figure~\ref{fig:feature_embedding}.
Sequentially, the estimated logits are coalesced by a product $\mathcal{F} = \prod\nolimits_{i = 1}^3 {{{\hat Y}_i}}$ and fed in into the final classification layer.


\section{Experimental Results}\label{experiments}
\begin{figure}[!hptb]
  \begin{center}
    \includegraphics[width=82mm]{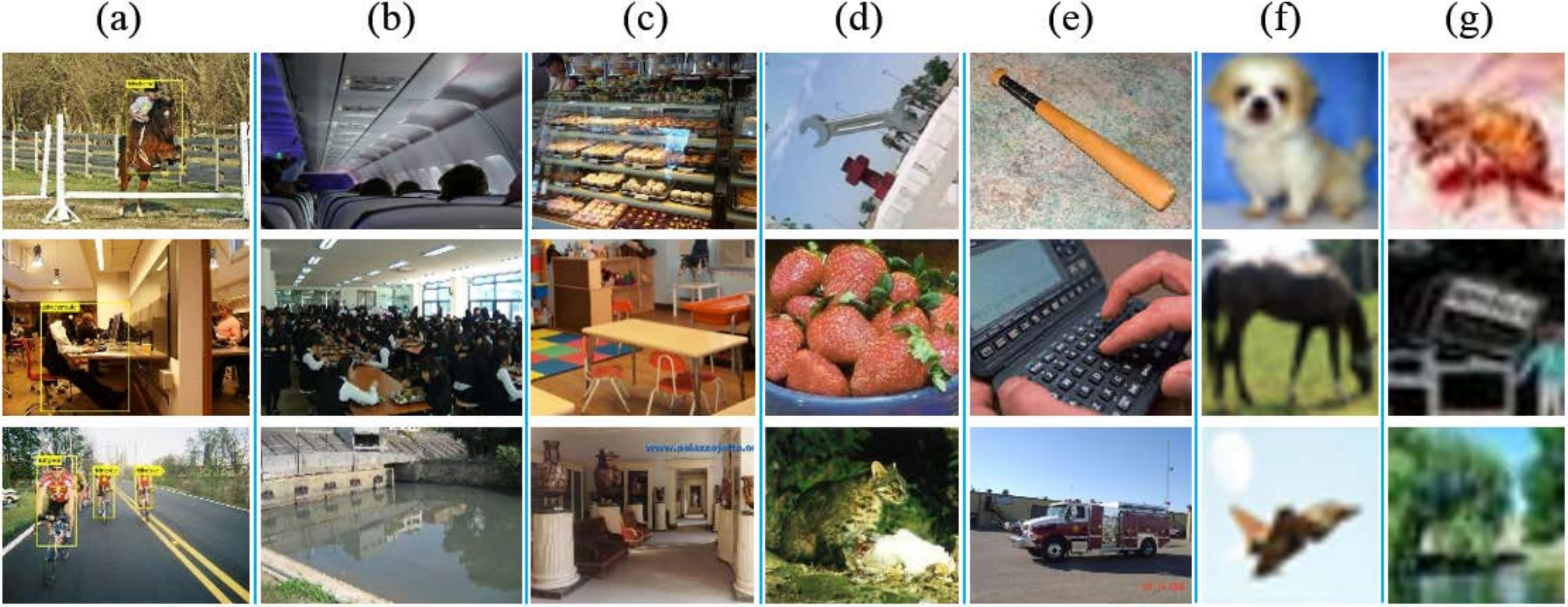}
  \end{center}
  \caption{Illustration of different data-set statistics: (a). Pascal VOC 2012 (riding horse, using computer, ridding bike), (b). Sun397 (airplane cabin, cafeteria, dam), (c). MIT67 (bakery, children room, museum), (d). Caltech101 (wrench, strawberry, wild cat), (e). Caltech256 (baseball bat, calculator, firetruck), (f). CIFAR10 (dog, horse, airplane), (g). CIFAR100 (insects, household furniture, large natural outdoor scenes).}
  \label{fig:samples}
\end{figure}

\begin{table*}[htp!]
\begin{center}
\begin{tabular}{|c||c|c|c|c|c|c|c|c|}
\hline
Type &Data-set & Proposed & AlexNet & VGG-16 & Ince.-v3 & Other methods \\
\hline\hline
\multirow{6}{*}{Object classification} & CIFAR10         & \textbf{92.00} & 81.60 & 85.35 & 89.57 &  91.87\cite{Sun16}, 85.02\cite{Snoek12}, 74.5\cite{Yu10}\\
                                       & CIFAR100        & \textbf{74.60} & 56.30 & 67.26 & 69.86 &  72.60\cite{icml2015_snoek15}, 66.64\cite{Sun16}\\
                                       & Caltech101      & \textbf{95.65} & 90.15 & 91.31 & 93.57 &  83.60\cite{Park10}, 82.10\cite{Gehler09}, 76.1\cite{Khan12}\\
                                       & Caltech256      & \textbf{87.30} & 69.22 & 79.30 & 83.75 &  60.97\cite{Dixit15}, 50.80\cite{Gehler09}\\
                                       & MIT67           & \textbf{77.38} & 53.88 & 66.41 & 76.04 &  70.72\cite{Zhou16}, 65.10\cite{Dixit15}\\
                                       & Sun397          & \textbf{55.22} & 45.18 & 47.87 & 49.41 &  54.30\cite{Zhou16}, 38.00\cite{Xiao10}\\
\hline
Action classification & Pascal VOC 2012 & \textbf{82.50} & 63.39 & 71.13 & 79.98 &  70.20\cite{Oquab:2014:LTM:2679600.2680210}, 69.60 OXFORD\cite{Everingham2015}\\
\hline
\end{tabular}
\end{center}
\caption{Comparison of the results (top-1 accuracy in \%).}
\label{table:results_comparison}
\end{table*}
Experiments were carried out on 6 different object classification data-sets: CIFAR-10, CIFAR-100 \cite{Krizhevsky09}, MIT67 \cite{Quattoni09} Caltech101, Caltech256 \footnote{\url{http://www.vision.caltech.edu/Image_Datasets/Caltech101/}}, Sun397 \footnote{\url{http://groups.csail.mit.edu/vision/SUN/}} and an action classification data-set the Pascal VOC 2012 \cite{Everingham2015}. Three statistics from each data-set is shown in Figure~\ref{fig:samples} while Table~\ref{table:data-sets} summarizes all the data-sets. In Pascal VOC 2012, as the action boundaries were given we extracted the action statistics within the boundaries and zero padded to make their dimension spatially square and resized to meet the requirement of the employed CNN architectures. For other data-sets, whole size images were taken and only resized to meet the networks' input layer requirements.

\begin{figure}[!htbp]
  \begin{center}
    \includegraphics[width=80mm]{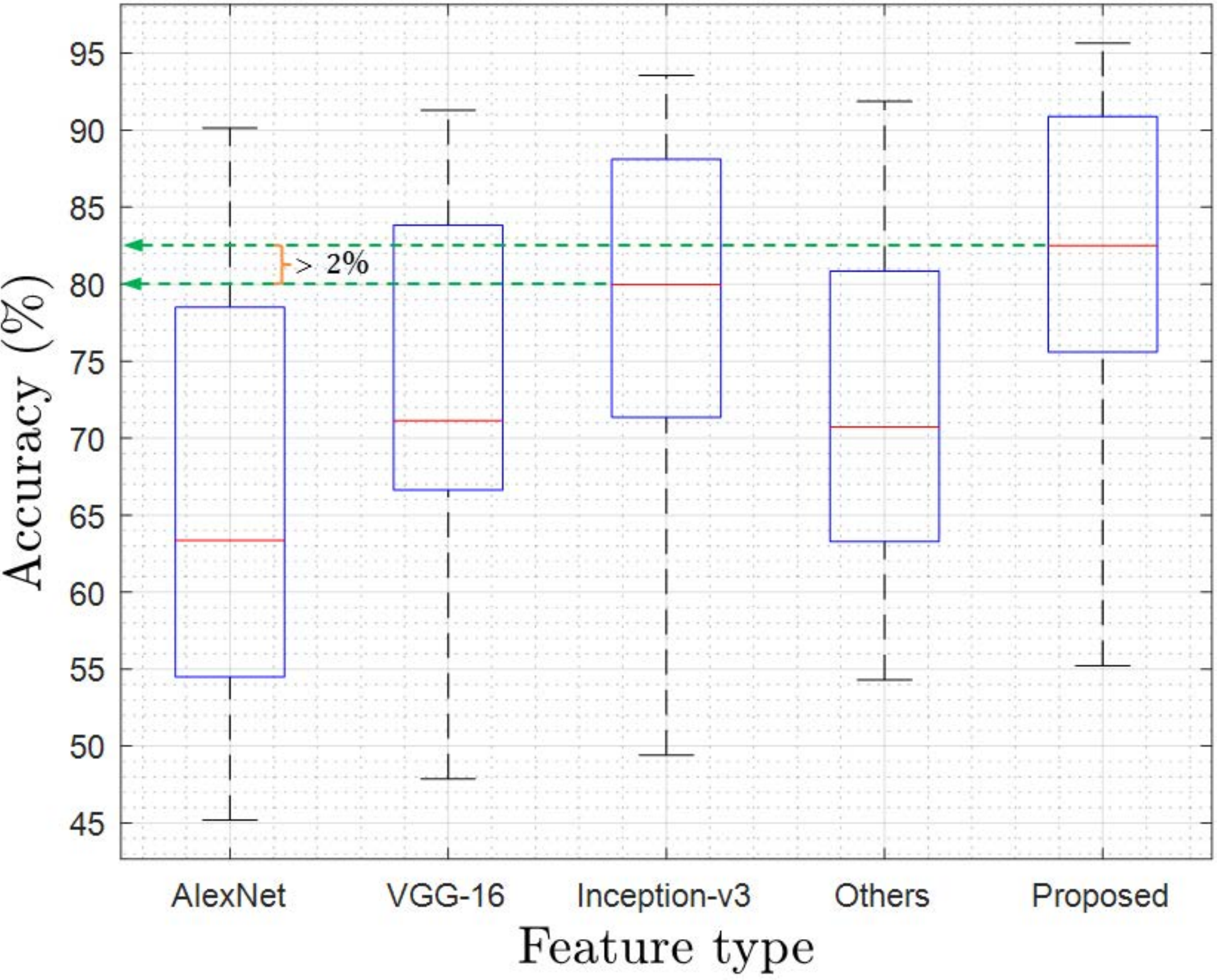}
  \end{center}
  \caption{Performance comparison.}
  \label{fig:Performance-comparison}
\end{figure}

The results of the proposed bottleneck feature embedding are compared in Table~\ref{table:results_comparison} with existing algorithms. The Table also lists the performance of single CNN bottleneck feature without any feature fusion for quantitative analysis, while Figure~\ref{fig:Performance-comparison} shows an overall performance comparison in terms of box-plot of the fused feature with the best results of other methods chosen from Table~\ref{table:results_comparison}. From these comparisons one can understand that the proposed feature embedding has improved the classification accuracy by 1\% - 2\% most of the cases without any data-augmentation.

\begin{table}[h!]
\begin{center}
\resizebox{\columnwidth}{!}{%
\begin{tabular}{|c|c|c|c|c|}
\hline
Data-set & No. of classes & Train. samples & Test samples & Ref. \\
\hline\hline
CIFAR10    & 10  & 50,000 & 10,000 & \cite{Krizhevsky09}\\
CIFAR100   & 100 & 50,000 & 10,000 & \cite{Krizhevsky09}\\
Caltech101 & 101 & 6,076  & 2,601  & \cite{Fei-Fei04}\\
Caltech256 & 256 & 21,363 & 9,146  & \cite{Griffin07}\\
MIT67      & 67  & 5,360  & 1,340  & \cite{Quattoni09}\\
Sun397     & 397 & 59,550 & 10,919 & \cite{Xiao10}\\
Pascal VOC & 10  & 4,588  & 4,569  & \cite{Everingham2015}\\
\hline
\end{tabular}
}
\end{center}
\caption{Summary of the data-sets.}
\label{table:data-sets}
\end{table}
Note that in Table~\ref{table:results_comparison}, \cite{Sun16} uses Data-augmentation + latent model ensemble with single CNN feature; \cite{Snoek12}, \cite{Yu10} and \cite{icml2015_snoek15} do not use any feature fusion; \cite{Park10}, \cite{Gehler09}, \cite{Khan12}, \cite{Dixit15} and \cite{Xiao10} use feature fusion of multiple hand-crafted features or hand-crafted feature(s) with a single CNN feature; \cite{Zhou16} uses CNN features extracted though pre-trained AlexNet on Places205/365, similarly \cite{Oquab:2014:LTM:2679600.2680210} also uses CNN features extracted by using a pre-trained AlexNet on 1512 classes of ImageNet (in our case, the AlexNet used is pre-trained on 1000 classes of ImageNet).


\section{Conclusion}\label{conclusion}
An approach to fuse bottleneck features of multiple CNNs through weighted cross-entropy is presented, where a set of three different pre-trained CNNs are exploited as feature extractors. The test results on various data-sets show that it outperforms the state-of-the-art hand-crafted feature fusion methods and produces very competitive results to fully trained (data-set specific) DCNN, as well. It accords with our hypothesis that features from multiple CNNs can be complementary to each other and fusion of them can be a generalized representation of images that is appearance invariant.

Although, the proposed feature embedding enhances the classification accuracy, how to fuse multiple features is still an open problem. In this work, our goal is to analyze if the accuracy improves when multiple CNN bottleneck features are fused as proposed. As for the future work, metric learning approaches can be exploited to capture facet in the CNN features that to differentiate classes and inter-classes. Hence, this work can be extended for dynamic texture and video activity detection and classification, as well.


\bibliography{egbib}
\bibliographystyle{ieeetr}


\end{document}